\documentclass{article} 
\usepackage{nips15submit_e,times}
\usepackage{hyperref}
\usepackage{url}

\usepackage{amsmath}
\usepackage{amsfonts}
\usepackage{amssymb}
\usepackage{mathtools}
\usepackage{graphicx}
\usepackage{caption}
\usepackage{subcaption}
\usepackage{verbatim}
\usepackage{algorithm}
\usepackage{algorithmic}

\title{Stochastic Collapsed Variational Inference for Hidden Markov Models}

\author{
Pengyu Wang$^1$\ \ \ \ \  Phil Blunsom$^{1,2}$\\
$^1$Department of Computer Science, University of Oxford\\
$^2$Google DeepMind \\
\texttt{\{pengyu.wang,\,phil.blunsom\}@cs.ox.ac.uk}
}

\nipsfinalcopy 

\begin{document}

\maketitle

\section{Introduction}

Hidden Markov models (HMMs) \cite{rabiner90hmm} are popular probabilistic models for modelling sequential data in a variety of fields including natural language processing, speech recognition, weather forecasting, financial prediction and bioinformatics. However, their traditional inference methods such as variational inference (VI) \cite{beal03} and Markov chain Monte Carlo (MCMC) \cite{scott02} are not readily scalable to large datasets. For example, one dataset in our experiment consists of $100$ million observations.

An important milestone for scaling VI was made by Hoffman et al.\ \cite{hoffman13}, who proposed stochastic VI (SVI) that computes cheap gradients based on minibatches of data, updating the model parameters before a complete pass of the full dataset. A recent scalable and more accurate algorithm was proposed by Foulds et al.\ \cite{foulds13}, who applied such stochastic optimization to the collapsed latent Dirichlet allocation (LDA) \cite{teh07collapsed}, and their stochastic collapsed variational inference (SCVI) algorithm has been successful in large scale topic modelling.

However, while these recent advances have been studied extensively for topic models that assume a simple bag-of-words data setting \cite{hoffman13,teh07collapsed,hoffman10,wang12,bryant12}, there has been little research on whether and how we can apply them in a time dependent data setting. Some research such as SVI for Bayesian time series models \cite{johnson14} and collapsed VI (CVI) for HMMs \cite{wangandblunsom13} consider the settings where datasets consist of many independent time series, naturally avoiding to break the sequential dependencies. Perhaps the only true exception is the SVI algorithm for HMMs proposed by Foti et al.\ \cite{foti14} in the setting of a single long time series, where the sequential dependencies must be broken.

In this paper, we follow the success of SCVI for LDA \cite{foulds13} and study the SCVI algorithm applied to a single long time series. In a collapsed HMM, we break a long chain into subchains, and we propose a novel sum-product algorithm to update the posteriors of subchains, taking into account their edge transitions due to the sequential dependencies. Our sum product algorithm can be understood as an alternative buffering method to the one in \cite{foti14}. Our experiments on two discrete datasets show that our SCVI algorithm for HMMs is scalable to very large datasets, memory efficient and significantly more accurate than the existing SVI algorithm.

\section{Background}

A hidden Markov model (HMM) \cite{rabiner90hmm} consists of a hidden state sequence $\textbf{z} = \{z_t\}_{t=0}^T$ and a corresponding observation sequence $\textbf{x} = \{x_t\}_{t=1}^T$. Let there be $K$ hidden states. For convenience, we let the start state be $0$ and set $z_0 = 0$. Let $\boldsymbol\theta$ be the transition matrix where $\theta_{k,k'}=p(z_t=k'|z_{t-1}=k)$, and $\theta_0$ be the initial state distribution where $\theta_{0,k'}=p(z_1=k')$. For $k=0,...,K$, we specify the Dirichlet priors with symmetric hyperparameters $\alpha$ on $\theta_k$, $\theta_k|\alpha \sim \text{Dir}(\alpha)$ in a Bayesian setting.

A hidden sequence is generated by a Markov process, and each observation is generated conditioned on its hidden state. We have for $t=1,...,T$,
\begin{align}
& z_t|z_{t-1}=k \sim \text{Mult}(\theta_{k}) &  x_t|z_t=k'  \sim p(\cdot|\phi_{k'}), &
\end{align}
where $\phi_{k'}$ parametrizes the observation likelihood for the hidden state $k'$, with $\phi_{k',w}=p(x_t=w|z_t=k')$. Without loss of generality, we assume that the observation likelihoods and their conjugate prior take exponential forms. The exponential family is a broad class of probability distributions including multinomial, Gaussian, gamma, Poisson, Dirichlet, Wishart and many others; and there is a conjugate prior distribution for each member in this class. We have for $k'=1,...,K$,
\begin{align}
 p(w|\phi_{k'}) &= h_l(w) \exp \{ \phi_{k'}^T t(w) - a_l(\phi_{k'}) \}  \\
 p(\phi_{k'}|\lambda^\circ) &= h_g(\phi_{k'}) \exp \{ (\lambda_1^\circ)^T \phi_{k'} + (\lambda_2^\circ)^T (-a_l(\phi_{k'})) - a_g(\lambda^\circ) \} .
\end{align}

The base measure $h$ and log normalizer $a$ are scalar functions; and the parameter $\phi_{k'}$ and sufficient statistics $t$ are vector functions. The subscripts $l$ and $g$ represent the local hidden variables and global model parameters, respectively. The dimensionality of the prior hyperparameter $\lambda^\circ = (\lambda_1^\circ, \lambda_2^\circ)$ is equal to $\text{dim}(\phi_{k'})+1$.

\section{Stochastic Collapsed Variational Inference}

There is substantial empirical evidence \cite{foulds13,wangandblunsom13,asuncion09} that marginalizing the model parameters is helpful for both accurate and efficient inference. Thus we integrate out the model parameters $(\boldsymbol\theta,\boldsymbol\phi)$ and the marginal data likelihood of an HMM is:
\begin{align}
 p(\textbf{x},\textbf{z}) = \prod_{k=0}^K \textstyle \frac{\Gamma(K\alpha)}{\Gamma(K\alpha+C_{k\cdot})} \prod_{k'=1}^K \frac{\Gamma(\alpha + C_{kk'})}{\Gamma(\alpha)}
 \displaystyle \prod_{t=1}^T \textstyle h_l(x_t)  \displaystyle \prod_{k'=1}^K \textstyle \exp \{ a_g(\lambda^{k'}) -  \{ a_g(\lambda^\circ)\}.
 \label{collapsedhdphmm}
\end{align}

The gamma functions and log normalizers result from the marginalization. $C_{kk'}$ denotes the transition count from the hidden state $k$ to $k'$, $C_{kk'} = \# \{t: z_{t-1}=k,z_t={k'}\}$. dot denotes the summed out column, e.g.,\ $C_{\cdot k'}= \sum_{k} C_{kk'}$. $\lambda^{k'}$ denotes the posterior hyperparameter for the hidden state $k'$, $\lambda^{k'}_1 = \lambda^\circ_1 + \sum_{t=1}^T t(x_t)\delta(z_t=k')$ and $\lambda^{k'}_2 =\lambda^\circ_2 + C_{\cdot k'}$, where $\delta$ is the standard delta function.

Given an observed sequence $\textbf{x}$, the task of Bayesian inference in the collapsed space is to compute the posterior distributions over the hidden sequence, $p(\textbf{z}|\textbf{x})$. The posteriors over the model parameters can be estimated by taking a variational Bayesian maximization step with our estimated $q(\textbf{z})$ \cite{beal03}. As the exact computation is intractable, we introduce a variational distribution $q(\textbf{z})$ in a tractable family and we maximize the evidence lower bound (ELBO) denoted by $\mathcal{L}(q)$,
\begin{align}
\log p(\textbf{x}) \geq \mathbb{E}[\log p(\textbf{x},\textbf{z})] - \mathbb{E}[\log q(\textbf{z}) ] \triangleq \mathcal{L}(q).
\end{align}

We consider the tractable family under the generalized mean field assumption \cite{xing02} in the collapsed space: we break a single long hidden sequence into a set of subchains. We have $q(\textbf{z}) = \prod_{n=1}^N q(\textbf{z}^n)$. We do not make any further assumptions about the inner structure of each subchain, preserving the inner transition information. It might be worth emphasizing that the time series dependencies in an HMM model are not broken; only the variational posterior is factorized. Therefore, the information can still flow across different subchains via edge transitions.

For notational simplicity, we let each subchain be of the length $L$ and $N=\left \lfloor {T/L}\right \rfloor$ be the number of subchains given a long chain. For each hidden subchain $\textbf{z}^n=\{z^n_l\}_{l=1}^L$, we denote the corresponding observed subchain by $\textbf{x}^n=\{x^n_l\}_{l=1}^L$. Combining the work of SCVI for LDA \cite{foulds13} and CVI for HMM \cite{wangandblunsom13}, we uniformly sample an observation subchain $\textbf{x}^n$, and we derive the posterior update for $q(\textbf{z}^n)$ with a zeroth order Taylor approximation \cite{teh07collapsed},
\begin{align}
 q(\textbf{z}^n)  &\approx \propto \hat{\theta}_{\cdot,z^n_1} \bigg(\prod_{l=2}^L \hat{\theta}_{z^n_{l-1},z^n_l} \bigg)  \hat{\theta}_{z^n_{L},\cdot} \bigg( \prod_{l=1}^L \hat{\phi}_{z^n_l,x^n_l} \bigg) \label{scvihmm1-1} \\
  \hat{\theta}_{\cdot,z^n_1}  &\propto  \sum_{z^n_0} q(z^n_0) \bigg(\mathbb{E}[C_{z^n_0,z^n_1}]+ \frac{\alpha}{Kq(z^n_0)} \bigg) \label{scvihmm1-2} \\
 \hat{\theta}_{z^n_{l-1},z^n_l}  &\propto \mathbb{E}[C_{z^n_{l-1}z^n_l}] + \alpha \label{scvihmm1-3} \\
 \hat{\theta}_{z^n_L,\cdot} &\propto \sum_{z^n_{L+1}} \bigg(\frac{\mathbb{E}[C_{z^n_L,z^n_{L+1}}]+\frac{\alpha}{Kq(z^n_{L+1})}}{\mathbb{E}[C_{z^n_L,\cdot}]+K\alpha}\bigg) q(z^n_{L+1}) \label{scvihmm1-4} \\
\hat{\phi}_{z^n_l,x^n_l} &\propto h(x^n_l) \exp \{ a_g( \lambda^\circ_1+t(x^n_l)+\mathbb{E}[t_{z^n_l}(\textbf{x},\textbf{z})],  \lambda^\circ_2+1+\mathbb{E}[C_{\cdot z^n_l}]) \},
\label{scvihmm1-5}
\end{align}
where $\mathbb{E}[C_{kk'}]=\sum_{t=1}^T q(z_{t-1},z_t=k,k')$ denotes the global expected transition count from state $k$ to $k'$, and $\mathbb{E}[t_{k'}(\textbf{x},\textbf{z})] = \sum_{t=1}^T q(z_t=k')t(x_t)$ denotes the global emission statistics at hidden state $k'$. Unlike CVI for HMM \cite{wangandblunsom13}, we do not need to maintain local statistics and thus our algorithm is memory efficient. We show the algorithmic procedure to infer $q(\textbf{z}^n)$ in Section 3.1.

Given $q(\textbf{z}^n)$, we can collect the local transition counts $\mathbb{E}[C^n_{kk'}]$ and emission statistics $\mathbb{E}[t_{k'}(\textbf{x}^n,\textbf{z}^n)]$ and update the global statistics with an online average weighted by a step size $\rho_n$,
\begin{align}
\mathbb{E}[C_{kk'}] &= (1-\rho_n) \mathbb{E}[C_{kk'}] + \rho_n T/(L-1) \mathbb{E}[C^n_{kk'}] \label{scvihmm2-1}\\
\mathbb{E}[t_{k'}(\textbf{x},\textbf{z})] &= (1-\rho_n) \mathbb{E}[t_{k'}(\textbf{x},\textbf{z})] + \rho_n N \mathbb{E}[t_{k'}(\textbf{x}^n,\textbf{z}^n)].
\label{scvihmm2-2}
\end{align}

\begin{figure*}
\centering
\includegraphics[scale=0.5]{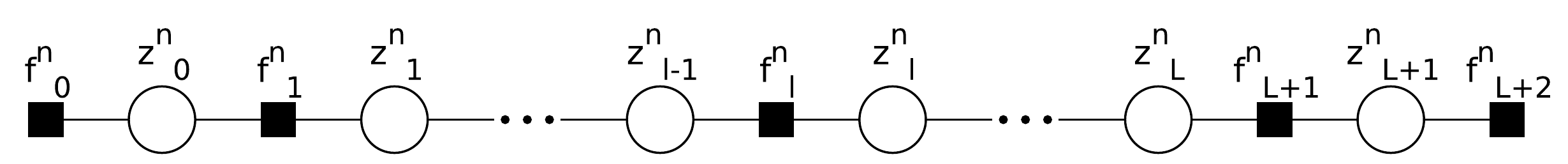}
\caption{The factor graph of a subchain $\textbf{z}^n=\{z^n_l\}_{l=1}^L$ and its guarding variables $z^n_0$ and $z^n_{L+1}$. The emission probabilities have been absorbed into the transition factors.}
\label{figure1}
\end{figure*}

\subsection{Modified Forward Backward Algorithm}

Given a subchain $\textbf{z}^n=\{z^n_l\}_{l=1}^L$, we denote the hidden variable before it $z^n_0$ and the hidden variable after it $z^n_{L+1}$ to be the guarding variables; and we denote $\hat{\theta}_{\cdot,z^n_1}$ and $\hat{\theta}_{z^n_{L},\cdot}$ to be the edge transitions. In (\ref{scvihmm1-1}), the edge transitions prevent us from applying the standard forward backward algorithm \cite{baum66} to the HMM parametrized by the surrogate parameters $\hat{\theta}$ and $\hat{\phi}$. Therefore, we propose a modified sum-product algorithm to buffer subchain edges with guarding variables. We start by defining a joint distribution of a subchain and its guarding variables using a factor graph shown in figure \ref{figure1},
\begin{align}
q(\textbf{z}^{(n)},z^n_0,z^n_{L+1}) &\propto f^n_0(z^n_0) \bigg(\prod_{l=1}^{L+1} f^n_l(z^n_{l-1},z^n_l)\bigg) f^n_{L+2}(z^n_{L+1}) \label{sumproduct-1} \\
f^n_0(z^n_0) &\triangleq q(z^n_0) \label{sumproduct-2} \\
f^n_1(z^n_0,z^n_1) &\triangleq \bigg(\mathbb{E}[C_{z^n_0,z^n_1}]+ \frac{\alpha}{Kq(z^n_0)}\bigg) \hat{\phi}_{z^n_1,x^n_1} \\
f^n_l(z^n_{l-1},z^n_l) &\triangleq  \hat{\theta}_{z^n_{l-1},z^n_l} \hat{\phi}_{z^n_l,x^n_l} \quad\quad\quad \text{for } l = 2,...,L  \\
f^n_{L+1}(z^n_L,z^n_{L+1}) &\triangleq \frac{\mathbb{E}[C_{z^n_L,z^n_{L+1}}]+\frac{\alpha}{Kq(z^n_{L+1})}}{\mathbb{E}[C_{z^n_L,\cdot}]+K\alpha} \\
f^n_{L+2}(z^n_{L+1}) &\triangleq q(z^n_{L+1}) \label{sumproduct-6}.
\end{align}

The functions associated with each factor node $\{f^n_l\}_{l=0}^{L+2}$ are given in (\ref{sumproduct-2}-\ref{sumproduct-6}). It is easy to verify that summing over the guarding variables of the joint probability in (\ref{sumproduct-1}) reduces to $q(\textbf{z}^n)$ in (\ref{scvihmm1-1}). Now we can use the sum product algorithm \cite{Kschischang06} to compute the required marginals of $q(\textbf{z}^n)$. Specifically, we first pick $f_{L+2}$ as the root node and pass the messages from the leaf node $f_0$, and then we pass messages in a reverse direction\footnote{In both recursions, we have eliminated the messages of the `variable node to factor node' type \cite{bishop06}.} . We have,
\begin{align}
 u_{f^n_l \rightarrow z^n_l}(z^n_l) &=  \sum_{z^n_{l-1}} u_{f^n_{l-1} \rightarrow z^n_{l-1}}(z^n_{l-1}) f^n_l(z^n_{l-1},z^n_l) \quad \text{ for } l=1,...,L+1 \\
 u_{f^n_{l+1}\rightarrow z^n_l}(z^n_l) &= \sum_{z^n_{l+1}} u_{f^n_{l+2}\rightarrow z^n_{l+1}}(z^n_{l+1}) f^n_{l+1}(z^n_l,z^n_{l+1}) \quad \text{ for } l=L,...,0,
\end{align}
where the initial messages are simply the distributions of the two guarding variables,
\begin{align}
& u_{f^n_0 \rightarrow z^n_0}(z^n_0) = f^n_0(z^n_0) & u_{f^n_{L+2} \rightarrow z^n_{L+1}}(z^n_{L+1}) = f^n_{L+2}(z^n_{L+1}).  & \label{initial}
\end{align}

After the messages have been passed in both directions, we compute the required variable marginals $q(z^n_l)$ and pairwise marginals $q(z^n_{l-1},z^n_{l})$ by,
\begin{align}
q(z^n_l) &\propto u_{f^n_l \rightarrow z^n_l}(z^n_l) u_{f^n_{l+1} \rightarrow z^n_l}(z^n_l) \\
q(z^n_{l-1},z^n_l) &\propto f^n_l(z^n_{l-1},z^n_l) u_{f^n_{l-2} \rightarrow z^n_{l-1}}(z^n_{l-1}) u_{f^n_{l+1} \rightarrow z^n_l}(z^n_l)
\end{align}

The normalization constant can be obtained by normalizing any of these marginals. This completes our algorithm to infer $q(\textbf{z}^n)$. 

Our modified sum product algorithm is an alternative buffering method to the one proposed by Foti et al.\ \cite{foti14} in their SVI algorithm for a single long time series. A key difference is that we assume the independent subchains and we allow messages to be passed across the boarders via local beliefs of the guarding variables in (\ref{initial}), whereas the subchains in the SVI algorithm are naturally correlated. However, the price for preserving the correlation is that they assume the hidden chain is irreducible and aperiodic so that each subchain starts with the initial distribution equal to the stationary distribution of the whole chain. A second superficial difference is that we buffer a subchain by only two guarding variables, whereas Foti et al.\ buffered a subchain with more observations.

\begin{figure*}[t]
\flushleft
\includegraphics[scale=0.365]{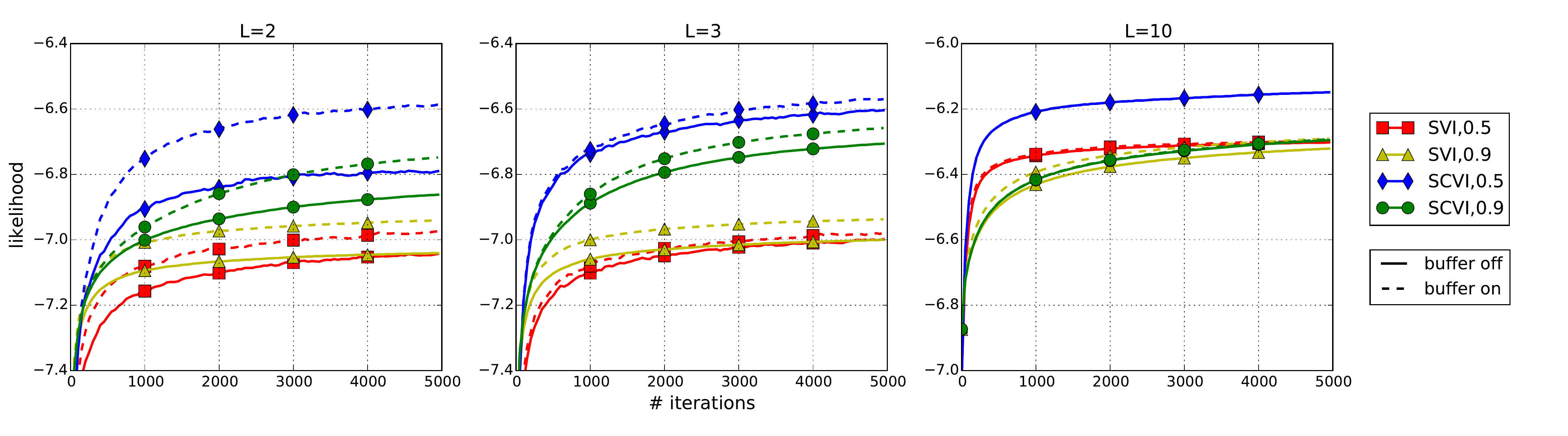}
\caption{Left and Middle: effect of incorporating buffering methods and performance comparison on WSJ. Right: performance comparison on NYT.}
\label{single_combined}
\end{figure*}

\section{Experiments}

We evaluated the utility of our buffering method and compared the performances of our SCVI algorithm against the SVI algorithm on two synthetic datasets created from the Wall Street Journal (WSJ) and New York Times (NYT). Both corpora are made of sentences, which in turn are sequences of words. For each sentence, the underlying sequence can be understood as a Markov chain of hidden part-of-speech (PoS) tags \cite{jurafsky00} and words are drawn conditioned on PoS tags, making HMMs natural models. We shuffled both datasets, added special symbols after each sentence to denote the ends and concatenated them. We used the first $1$ million words in the concatenated WSJ and $100$ million words in the concatenated NYT as our two long time series, respectively. As the evaluation metrics, we used predictive log likelihoods by holding out $5\%$ words of each time series as testing sets.

For both the SVI and our SCVI algorithms: we set the transition and emission priors to be $\text{Dir}(0.1)$; we initialized the global statistics using exponential distributions suggested by Hoffman et al.\ \cite{hoffman13}; we set $K=12$ assuming a universal PoS tag set \cite{petrov12}; when buffering was turned off, we set the initial distribution to start a subchain to be the whole chain's stationary distribution. For SVI, when buffering was turned on, we buffered a subchain with $20$ words on both sides. We varied the subchain lengths, $L=2,3,10$ and used minibatches of subchains to reduce the sampling variance. Following Foti et al.\ \cite{foti14}, we fixed the total length of all subchains in a minibatch $L \times M = 1000$, where $M$ is the minibatch size. Increasing $L$ means decreasing $M$ and vice-versa. Also, we varied the forgetting rates $\kappa=0.5,0.9$, which parametrize the step sizes $\rho_n = (1+n)^{-\kappa}$. Under each of the combined settings, we ran both algorithms for $5000$ iterations.

Figure \ref{single_combined} presents the predictive log likelihood results on the WSJ (left and middle) and NYT (right). We see that in most settings our SCVI algorithm outperformed the SVI algorithm by large margins, extending the success of SCVI for LDA \cite{foulds13} to time series data. The only exception is when $\kappa=0.9$, both algorithms performed comparably on the NYT. For SCVI, a smaller forgetting rate was preferred, which further promotes the scalability; whereas SVI was less sensitive. When $L$ is small, there are noticeable improvements using respective buffering methods in both algorithms. For SCVI, we attribute the improvement to the inter subchain communication through guarding variables.

\section{Conclusion}

We have presented a stochastic collapsed variational inference algorithm for HMMs in the setting of a single long time series and an alternative buffering method that modifies the standard forward backward recursions. Our SCVI algorithm is significantly more accurate than the SVI algorithm on two large datasets, and our buffering method is robust against the poor choices of subchain lengths. For future work, we aim to derive the true nature gradients of the ELBO to prove the convergence of our algorithm \cite{ruiz2014}, although we never saw a nonconverging case in our experiments.

\small{
\bibliographystyle{unsrt}
\bibliography{standard}
}

\end{document}